\documentclass[letterpaper, 10 pt, conference]{ieeeconf}  

\IEEEoverridecommandlockouts                              

\pdfoutput=1

\usepackage{algorithm}
\usepackage{algpseudocode}
\usepackage{graphicx}
\usepackage{subcaption}
\usepackage{tabularray}
\usepackage{tabularx}
\usepackage[bookmarks=true]{hyperref}
\UseTblrLibrary{diagbox}
\usepackage{float}
\usepackage{cite}
\usepackage{booktabs}  
\usepackage{amsmath}   
\usepackage{array}     
\usepackage{tensor}

\title{\LARGE \bf
\textbf{Planning by Simulation: Motion Planning with Learning-based Parallel Scenario Prediction for Autonomous Driving}
}

\author{Tian Niu, Kaizhao Zhang, Zhongxue Gan and Wenchao Ding$^{*}$ 
\thanks{*Corresponding author
}
\thanks{All authors are with Academy of Enigeering and Technology, Fudan University, Shanghai.
Email:
      {\tt\small \{tniu21, kzzhang24\}@m.fudan.edu.cn, \{ganzhongxue, dingwenchao\}@fudan.edu.cn}}%
}

\begin{document}

\maketitle
 \thispagestyle{empty}
\pagestyle{empty}

\begin{abstract}

Planning safe trajectories for autonomous vehicles is essential for operational safety but remains extremely challenging due to the complex interactions among traffic participants. Recent autonomous driving frameworks have focused on improving prediction accuracy to explicitly model these interactions. However, some methods overlook the significant influence of the ego vehicle's planning on the possible trajectories of other agents, which can alter prediction accuracy and lead to unsafe planning decisions. In this paper, we propose a novel motion Planning approach by Simulation with learning-based parallel scenario prediction (PS). PS deduces predictions iteratively based on Monte Carlo Tree Search (MCTS), jointly inferring scenarios that cooperate with the ego vehicle's planning set. Our method simulates possible scenes and calculates their costs after the ego vehicle executes potential actions. To balance and prune unreasonable actions and scenarios, we adopt MCTS as the foundation to explore possible future interactions encoded within the prediction network. Moreover, the query-centric trajectory prediction streamlines our scene generation, enabling a sophisticated framework that captures the mutual influence between other agents' predictions and the ego vehicle's planning. We evaluate our framework on the Argoverse 2 dataset, and the results demonstrate that our approach effectively achieves parallel ego vehicle planning.

\end{abstract}

\section{INTRODUCTION}
\begin{figure}[t] 
    \centering 
    \includegraphics[width=0.50\textwidth]{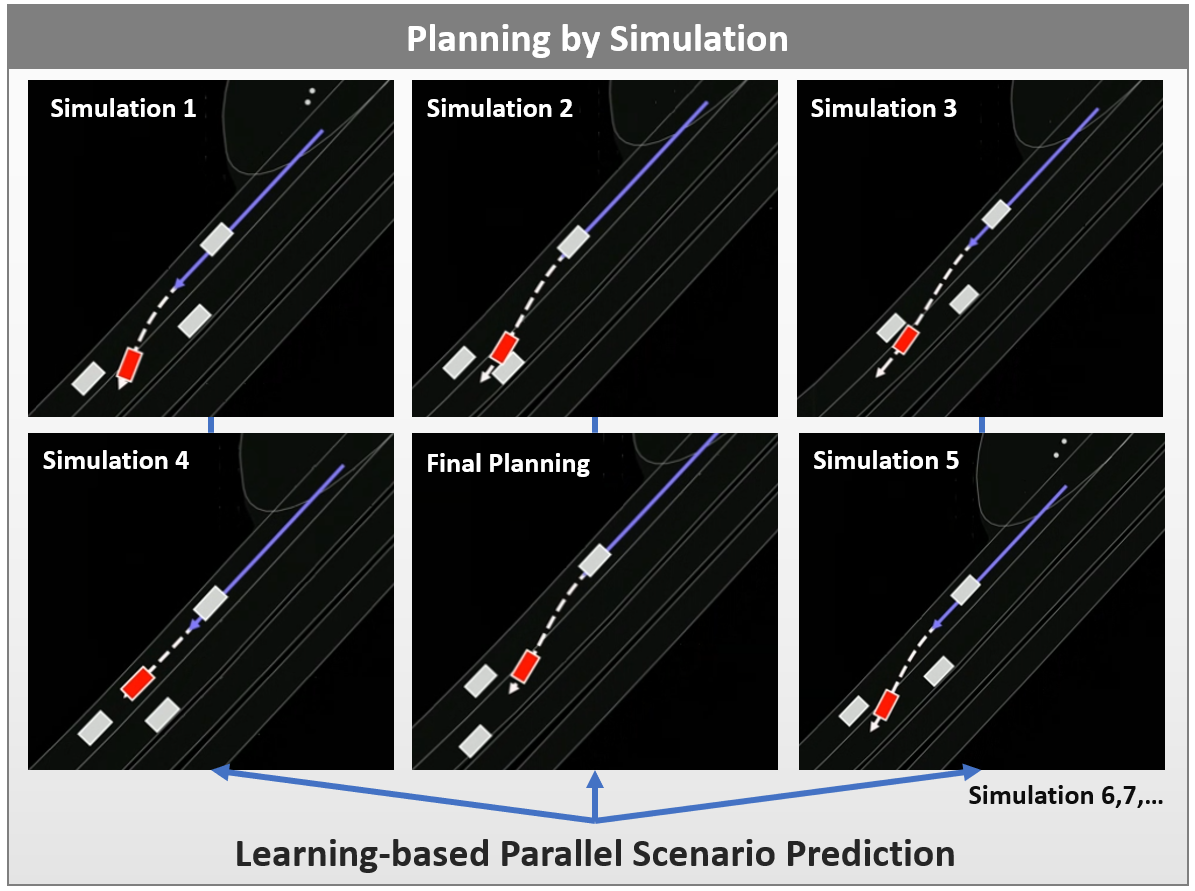} 
    \caption{Illustration of the planning by simulation. First, multiple future scenarios are predicted by our learning-based parallel prediction module, which simulates the ego vehicle and our vehicle simultaneously. The scenario prediction is conducted by a MCTS search. Safe and smooth ego planning are chosen out of multiple possible futures. Compared to pure e2e planning, our framework is more robust thanks to explicit future reasoning.
    More examples can be found in the video \href{https://youtu.be/_Y9HGzgKfsc}{https://youtu.be/\_Y9HGzgKfsc}.} 
    \label{page1} 
\end{figure}
\begin{figure*}[htbp] 
    \centering 
    \includegraphics[width=\textwidth]{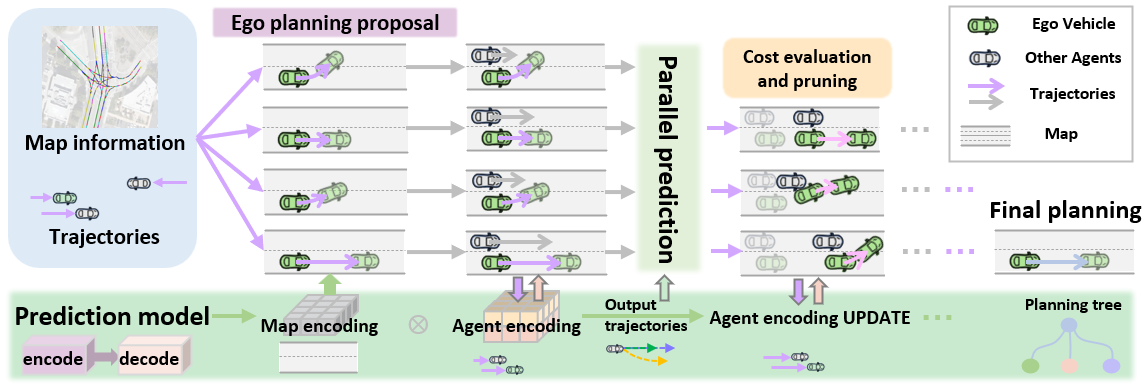} 
    \caption{Illustration of the proposed trajectory planning framework. Our model takes high-definition maps and agents' historical states as inputs to extract lane polygons and agents' trajectories. Based on these inputs, the model generates ego planning set. The embedded prediction model separates map encoding from agent encoding and outputs high-accuracy future trajectories for the agents. We execute the prediction process in parallel, expanding the search tree layer by layer. After that, the Monte Carlo policy helps prune branches with poor states, and the remaining states become new nodes, quickly encoded into the prediction model. This process repeats until the specified depth is reached, achieving a cyclic interaction between prediction and planning.} 
    \label{Pipeline} 
\end{figure*}

In autonomous driving, the fundamental capacity is to make efficient, safe, and human-like decisions. Both interpretable~\cite{guan2022integrated,brewitt2021grit} and inexplicable~\cite{bachute2021autonomous,van2018autonomous} methods have emerged in research on how to make better decisions. 
A growing trend of leveraging large-scale data in end-to-end learning-based planning has been shown in recent years~\cite{chen2024end}. End-to-end autonomous driving systems are fully differentiable programs that take raw sensor data as input and produce a decision as output~\cite{gu2023vip3d,sun2021complementing}. This jointly optimized solution has shown actual improvements in driving~\cite{ryan2020end}. However, uncertainty from surrounding vehicles and pedestrians may significantly affect the robustness of end-to-end systems~\cite{gonzalez2015review,hang2020integrated}.

Recently, large language model has sprung up and demonstrates abilities to empower autonomous driving in logical reasoning and generating answers~\cite{yang2023llm4drive}. OpenAI o1, the latest large language model, achieves complicated logical inference in high-level physics and coding problems. Unlike the early large language model, OpenAI o1 embeds a long internal chain of thought after pre-training and post-training, which helps replanning and self-modification to improve robustness. Inspired by these thoughts, we propose a new end-to-end framework like OpenAI o1, which conducts robust search based on end-to-end predictions. Different from directly outputting a single trajectory from end-to-end models, we use learned-based scenario prediction as a parallelizable simulator and plan by choosing the best out of different possible futures, as shown in \autoref{page1}.

Prediction provides possible trajectories of vehicles and pedestrians for downstream decisions~\cite{zhang2022ai,liu2022interaction}. Note that the prediction in this paper is \textit{scenario oriented}, which means that not only other traffic participants but also the ego vehicle (e2e planning) are predicted simultaneously.

In view of the multi-modality of prediction, not all planning modules are suitable for combining with prediction. Monte Carlo tree search (MCTS), an intelligent tree search method that balances exploration and exploitation~\cite{silver2010monte}, has a natural advantage on unfolding parallel circumstances of multiple possible decisions, which dramatically helps evaluate cost~\cite{chen2020driving,albrecht2021interpretable}. Some works propose a deep-learning heuristic MCTS algorithm to simulate surrounding space during planning~\cite{lei2021kb,weingertner2020monte}. 
Although they have achieved competitive results, there remain two main drawbacks: (1) Preordered nodes are repeatedly used while predicting the latest circumstance. These models fail to reuse past computations and operate streamingly. (2) These methods define discrete yaw rates and accelerated speed as action sets. A finite discrete action set may cause the planned trajectory to have a nonnegligible gap with the actual trajectory. Besides, the expansion of the search tree may generate an unreasonable branch that contradicts the laws of physics. 

Most prediction networks combined with search-based planning are agent-centric modeling scheme. They require re-normalizing and re-encoding the input whenever the observation window slides forward, leading to redundant computations if they are directly and repeatedly invoked~\cite{jia2022multi}. Query-centric trajectory prediction, different from agent-centric prediction, enables the reuse of past computations by learning representations independent of the global spacetime coordinate system~\cite{zhou2023query}. Sharing the invariant scene features among all target agents in the Monte Carlo search tree further allows the parallelism of simulating tree-like future. 

Motivated by these observations,  we propose an iterative motion Planning by Simulation through learning-based parallel scenario prediction (PS) for autonomous driving based on MCTS, as shown in \autoref{Pipeline}. Our model first generates an optional action set and then cooperates ego vehicle actions and other agents' future trajectories to ratiocinate future scenarios in parallel. The major contributions of this paper are summarized as:
\begin{itemize}
    \item  A planning framework with data-driven interaction-aware inference capabilities, which is more robust than pure learning-based planning and more scalable than rule-based counterpart.
    \item An iterative planning structure guided by trajectory prediction, which in turn influences the prediction results.
    \item A reliable and efficient planning set generation method, along with a reusable prediction network based on coordinate transformation. Comprehensive experiments and comparisons are conducted to validate the performance.
\end{itemize}
The related literature is reviewed in \autoref{sec:related work}. Our method is detailed in \autoref{sec:method}. Experimental results are elaborated in \autoref{sec:experiments}. Finally, a conclusion is drawn in \autoref{sec:conclusion}.

\section{RELATED WORK}
\label{sec:related work}
\subsection{MCTS-based Planning }

MCTS is a searching algorithm that combines the classical tree search and reinforcement learning. AlphaGo and AlphaZero have defeated human world champions by combining an MCTS with deep reinforcement learning~\cite{silver2017mastering}. Originating from similar game theory and artificial intelligence, MCTS also has applications in various autonomous fields, showcasing its adaptability and robustness. It is suitable for various tasks in multi-robot active perception~\cite{best2019dec,sukkar2019multi}. In autonomous driving control, MCTS has been leveraged to enhance feedback steering controllers for autonomous vehicles~\cite{tian2022enhancing}. By adapting MCTS to behavior planning, some works harness their intrinsic ability to balance exploration and exploitation, making it well-suited to the intricate, and uncertain nature of real-world traffic scenarios~\cite{albrecht2021interpretable,wen2023monte}.  
\begin{figure*}[htbp] 
    \centering 
    \includegraphics[width=\textwidth]{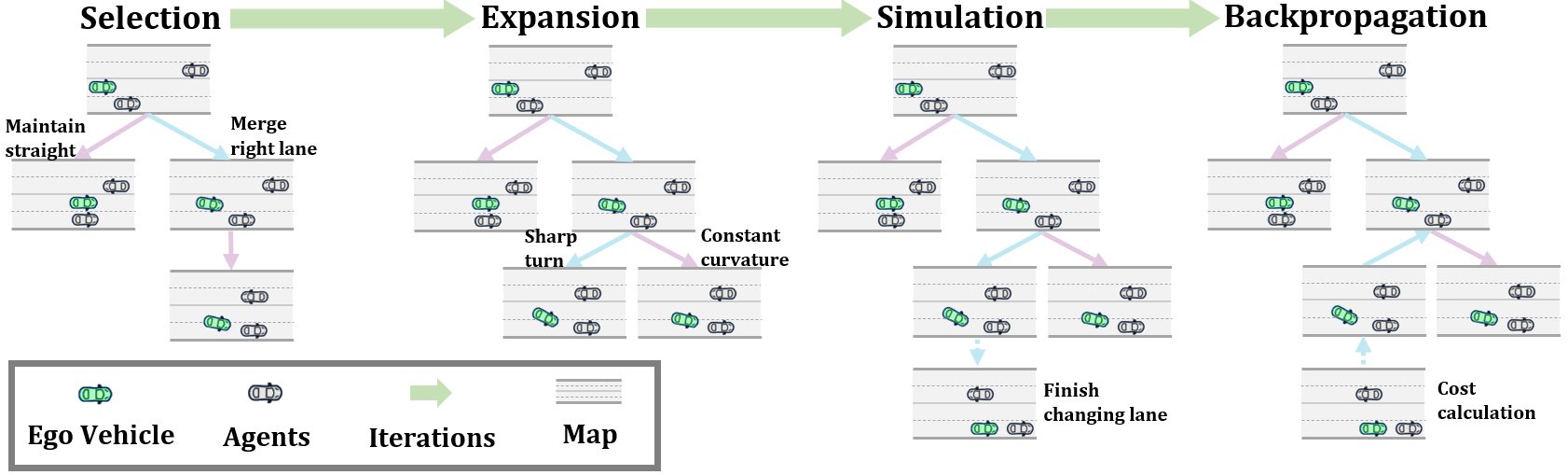} 
    \caption{Illustration of planning future scenarios. This reasoning process, which iterates dozens of times, is based on MCTS and is composed of four steps: selection, expansion, simulation, and backpropagation.} 
    \label{Tree Search Process} 
    \vspace{-0.5cm}
\end{figure*}
\subsection{Frenét Frame Trajectory Generation}

Rather than formulating the trajectory generation problem directly in Cartesian Coordinates, the Frenét frame method switches to find the best lateral and longitudinal acceleration changes along the dynamic reference lane~\cite{werling2010optimal}. Some works augment decision-making with Frenét frame path generator, improving the pedestrian handling ability by predicting their behavior~\cite{sarcinelli2019handling}. Other works project the semantic elements to the Frenét frame representation to provide a constraint-satisfied trajectory~\cite{ding2019safe,ding2021epsilon}. In addition, Frenét frame-based planners also perform well at high-speed driving~scenarios~\cite{raji2022motion}. However, few works incorporate the Frenét frame with a tree-like planner.

\subsection{Query-centric Trajectory Prediction }

A critical design in trajectory prediction is the reference coordinate system in which the representation is encoded~\cite{ngiam2021scene}, which would influence efficiency connected with planning. Scene-centric coordinate frame, sharing representation of world state, may sacrifice pose-invariance~\cite{cui2019multimodal}. Agent-coordinate frame, which is intrinsically pose-invariant, scales linearly with the number of agents~\cite{mercat2020multi}. They do not support being partially tested when a small part of historical trajectories have changed. Query-centric methods, encoding relative information, avoid the above problems and are suitable for tree search-based planners~\cite {zhou2023query}. This frame can also be used in end-to-end autonomous driving, where the sparse queries completely represent the whole driving scenario across space and time\cite{zhang2024sparsead}. 

\subsection{Interactive Prediction and Planning}

In classical autonomous vehicle tasks, separating prediction and planning layers limits the planner to reacting to predictions uninformed by the ego planned trajectory~\cite{espinoza2022deep}. Some works couple these layers via game theory that uses a novel interactive multi-agent neural network policy as part of its predictive model~\cite{huang2023gameformer,bahram2015game}. DTPP~\cite{huang2024dtpp} introduces a query-centric Transformer model that performs efficient ego-conditioned motion prediction, which guides tree-simulation planner to be pruned and expanded.

Our method is different from previous works. We introduce MCTS as a comprehensive decision-making framework for behavior planning, integrating Frenét frame to generate driving actions, which achieves layer-by-layer development of future scenarios. Besides, we utilize query-centric encoding framework to predict new nodes in our planner. 

\section{METHOD}
\label{sec:method}

The proposed framework with its associated prediction model is illustrated in \autoref{Pipeline}. Our planner is based on top of MCTS, with node pruning enabled by Frenét frame and cost evaluation. Besides, deep integration with trajectory prediction model significantly improves planning performance and efficiency.  
The key idea behind tree-structured planning is to approximate the intractable continuous-space policy planning problem by sampling a discrete set of ego trajectories in multiple stages. This structure forms a trajectory tree. Together with predicting the motion of other agents conditioned on each ego trajectory segment, we get a scenario tree. Given a sequence of past observed states $\left \{ s^{t}_{1},s^{t}_{2},...,s^{t}_{N}  \right \} _{t=1}^{T} $, and high-definition map information $M$, we aim to choose the most appropriate action $a\in A$ for ego agent to carry out at time $T+1$ that minimize cost $C$. $N$ represents the number of agents in the scene. We set $ \left \{ s^{t}_{1}\right \} _{t=1}^{T} $ as the state of the ego vehicle.
\floatname{algorithm}{Algorithm}
\renewcommand{\algorithmicrequire}{\textbf{Input:}}
\renewcommand{\algorithmicensure}{\textbf{Output:}}
\begin{algorithm}[b]
    
    \caption{Planning Process}\label{alg1}
    \begin{algorithmic}[1] 
        \Require sequence of past observed states $\{ s^{t}_{1},s^{t}_{2},...,s^{t}_{N}\}_{t=1}^{T} $ and high-definition map information $M$
        \Ensure the most appropriate action $a$
        \Function{Select Action}{$s_{n=1,...,N} ^{T},d$}
            \While{$iter<iterations$}
            \State \textsc{Simulate}($s_{n=1,...,N} ^{T},d$)
            \EndWhile
                \State \Return{$\mathrm{argmax} _{a} Q(s_{n=1,...,N} ^{T},a)$}
        \EndFunction
    \end{algorithmic}
\end{algorithm}
\subsection{Planning Process} 

Our proposed planning algorithm is outlined in Algorithm \ref{alg1}. We set $\left \{ s_{n} ^{i} \right \} _{n=1}^{N} $ ,$i\ge T$ as node states in the tree, amongst $s_{n} ^{i} = \left ( x_{n} ^{i},y_{n} ^{i},v\left ( x \right )  _{n} ^{i},v\left ( y \right ) _{n} ^{i},\theta _{n} ^{i} \right ) $, separately representing the $x$ axis and $y$ axis position, $x$ axis and $y$ axis velocity and steering angle. Different from normal binary action setting in autonomous driving, we define action $A=\left \{ 0.5,1.5,...,13.5,14.5 \right \} $ as the target speed set calculated for Frenét frame, which further generates paths in accordance with the law of kinematics alongside road center line. 


\begin{algorithm}
    \caption{Simulation Process}\label{alg2}
    \begin{algorithmic}[1] 
        \Function{Simulate}{$s,d$}
        \If{$d=0$}
        \State \Return{$0$}
        \EndIf
        \If{$s\notin T$}
        \For{$a\in A$}
        \State $Q(s,a),N(s,a)\gets 0,0$
        \EndFor
        \State $T=T\cup \left \{ s \right \} $
        \State \Return \textsc{Rollout}($s,d$)
        \EndIf
        \State $a\gets \mathrm{argmax} _{a} Q\left ( s,a \right ) +c\sqrt{\frac{\log{N(s)} }{N(s,a)} }$
        \State $s'\sim $ \textsc{Transition}($s,a$)
        \State $r\sim -$\textsc{Cost}($s,a$)
        \State $q\gets r+\lambda $\textsc{Simulate}($s',d-1$)
        \State $N(s,a)\gets N(s,a)+1$
        \State $Q(s,a)\gets Q(s,a)+\frac{q-Q(s,a)}{N(s,a)}$
        \State \Return $q$
        \EndFunction
        \State
        \Function{Rollout}{$s,d$}
        \If{$d=0$}
        \State \Return{$0$}
        \Else
        \State $a\sim $\textsc{Random}$(A)$
                \State $s'\sim $ \textsc{Transition}($s,a$)
                \State $r\sim -$\textsc{Cost}($s,a$)
                \State \Return $r+\lambda $\textsc{Rollout}($s',d-1$)
        \EndIf
        \EndFunction
    \end{algorithmic}
\end{algorithm}
As outlined in Algorithm \autoref{alg2} and as shown in \autoref{Tree Search Process}, The basic steps of our planner are similar to traditional MCTS~\cite{weingertner2020monte}. First, the selection step chooses a node that maximizes the upper confidence bound (UCB)  near the current node. UCB helps balance exploration and exploitation. Once reaching a state that is not part of the explored set, new leaves are expanded by iterating over all possible actions at the expansion step. After that, many random simulations are performed to a fixed depth to evaluate the value of the leaves at the rollout step. We simulate the final scene from the leaf node by setting the ego vehicle's random action. Finally, the statistics of all selected nodes are updated through backpropagation. With these four steps, our planner generates a growing asymmetric tree through continuous iterations until we meet a maximum number of simulations and the car has reached the goal point. Then we evaluate the performance and execute the action with maximum reward.

During the search, at selection step, we execute the action $a\in A$ that maximizes  $Q\left ( s,a \right ) +c\sqrt{\frac{\log{N(s)} }{N(s,a)} } $~\cite{auer2002finite}, where $N\left ( s\right )$ and $N\left ( s,a \right )$ track the number of times that a state and state-action pair are visited. Here $c$ is a hyper-parameter controlling the amount of exploration in the search. It encourages exploring less visited $\left ( s,a \right )$ pairs and relies on $Q\left ( s,a \right ) $ to estimate well explored pairs.

Our mean novelty is how to create child nodes given a target speed. Instead of using a single kinematic model, we update the ego vehicle state through the optimal Frenét frame method and update other agents by jointly predicting using selected nodes under the scene. 
\vspace{-0.08cm}
\subsection{Ego Vehicle Transition}
\label{sec:ego trans}

\begin{figure}[t] 
   \centering
  \begin{minipage}[b]{0.23\textwidth}
    \centering
    \includegraphics[width=\textwidth]{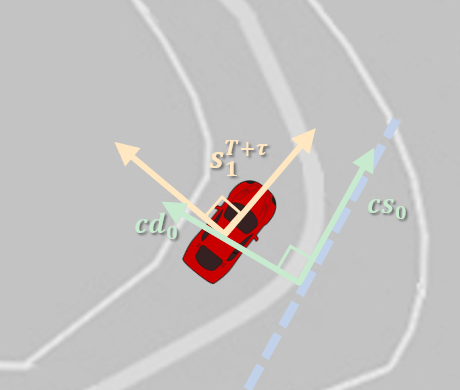}
    \caption*{(a)}
  \end{minipage}%
  \hspace{0.01\textwidth}
  \begin{minipage}[b]{0.23\textwidth}
    \centering
    \includegraphics[width=\textwidth]{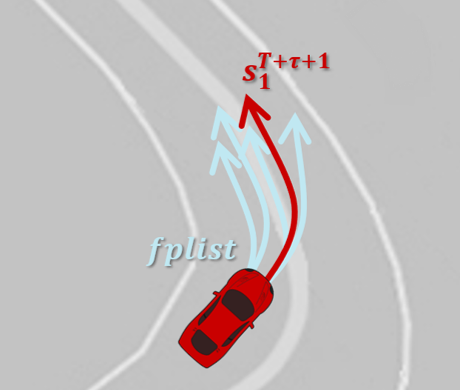}
    \caption*{(b)}
  \end{minipage}%
  
  \caption{Illustration of generating planning set for ego vehicle. (a) represents the coordinate frame relative to the centerline. (b) represents the generated path list and best path based on the transferred coordinate frame}
  \label{Frenet}
  \vspace{-0.2cm}
\end{figure}

Given high-definition map information $M$ and the target global route, we can calculate the directional vector of the target road center lines in the Cartesian coordinate system. 
The newest leaf node $s_{1} ^{T+\tau+1} = \left ( x_{1} ^{T+\tau+1},y_{1} ^{T+\tau+1},v\left ( x \right )  _{1} ^{T+\tau+1},v\left ( y \right ) _{1} ^{T+\tau+1},\theta _{1} ^{T+\tau+1} \right )$ on the branch are transformed from Cartesian coordinates to Frenét coordinates, through which we get the longitudinal and lateral position $\left ( cs_{0},cd_{0}   \right ) $ of ego vehicle relative to the center line. We then sample target extension time $\Delta t$ in the range $\left [ T_{min} , T_{max} \right ] $ to get longitudinal distance set $cT$ from the center line. To get homologous lateral position $cD$, we sample lateral displacement $\Delta d$ in the range $\left [ -m_{road,max} , -m_{road,max} \right ] $. Together with the chosen target speed $a\in A$, we generate the corresponding set of trajectories through quintic and quartic polynomial, denoted as $fplist = Polynomial\left ( cs_{0},cd_{0}\mid \Delta t\in cT ,\Delta d\in cD ,a \right ) $. By minimizing the cost function that is compliant with Bellman's principle of optimality, we choose the best path $bestpath\in fplist$. $Bestpath$ makes the best compromise between the jerk and the time. By inverse compute $bestpath$ in Cartesian coordinates, we get child node $s_{1} ^{i+1}$ of $s_{1} ^{i}$. The process is visualized in \autoref{Frenet}.

\subsection{Other Agents Transition}

\begin{figure}[t] 
    \centering 
    \includegraphics[width=0.5\textwidth]{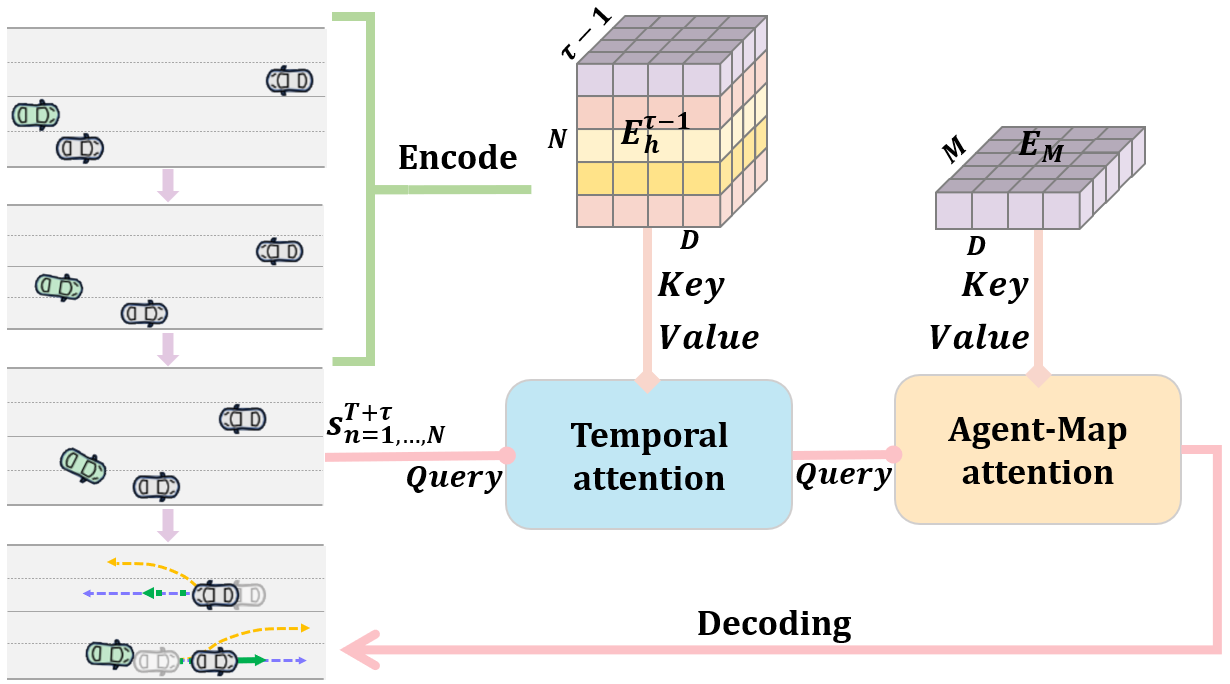} 
    \caption{Illustration of how to update other agents' states. Our model encodes history nodes on the branch as $\left [ N,\tau -1,D \right ] $ dimension agents embedding and $ \left [M ,D \right ] $ dimension map embedding, where $D$ represents the dimension of encoding hidden layer. When a new node $s_{n=1,...,N} ^{T+\tau }$ expands, temporal attention encodes $s_{n=1,...,N} ^{T+\tau }$  as query and historical agents encoding as key and value. After decoding, the model outputs $k$ future trajectories and their probabilities.} 
    \label{Other Agents Transition} 
    \vspace{-0.5cm}
\end{figure}

If independently calculating other agents' future trajectories, we would ignore the interaction from ego vehicle planning~\cite{zhang2022ai}. Assuming that the branch being updated is $h_{N} ^{\tau }=\left [ s_{n=1,...,N} ^{T},s_{n=1,...,N} ^{T+1},...,s_{n=1,...,N} ^{T+\tau } \right ] $, from the root node to leaf node, we have already get ego vehicle state $s_{1} ^{T+\tau+1 } $ in \autoref{sec:ego trans}. Our prediction model first encodes the road polyline information $M$ and history trajectory information $h_{N} ^{\tau }$ for all agents, denoted as $E^{\tau }$. Then, we decode $E^{\tau }$ into multi-modal trajectories and likelihoods. Normally, when new state $s_{n=1,...,N} ^{T+\tau }$ join in branch $h_{N} ^{\tau-1 }=\left [ s_{n=1,...,N} ^{T},s_{n=1,...,N} ^{T+1},...,s_{n=1,...,N} ^{T+\tau -1} \right ] $, we should re-encode $h_{N} ^{\tau }= h_{N} ^{\tau-1 }\cup s_{n=1,...,N} ^{T+\tau }$. Nevertheless, by transferring the Cartesian frame to a query-centric frame through relative spatial-temporal position calculation, we separately encode $h_{N} ^{\tau-1 }$ and $M$ into $E_{h} ^{\tau-1 }$ and $E_{M} $. Then $E^{\tau-1 }=SocialAttention\left ( E_{h} ^{\tau-1 },E_{M} \right ) $. By treating $s_{n=1,...,N} ^{T+\tau }$ as query of attention operator, $E_{h} ^{\tau }$ can be encoded temporally through $E_{h} ^{\tau-1 }$ and $s_{n=1,...,N} ^{T+\tau }$ directly. Because all these inference happens under the same scene, meaning $E_{M}$ stays the same, we can reuse the previous encodings $E^{\tau }=E^{\tau-1 }\oplus s_{n=1,...,N} ^{T+\tau }$. As shown in \autoref{Other Agents Transition}, this accurate and lightweight method predicts future positions $\left \{ \left ( \tensor*[]{^{k} x}{^{\tau+1}},\tensor*[]{^{k} y}{^{\tau+1}}  \right ),...,\left ( \tensor*[]{^{k} x}{^{\tau+1+\tau'}},\tensor*[]{^{k} y}{^{\tau+1+\tau'}}  \right )  \right \}_{k=1}^{6}  $ with probability $p_{1},...,p_{6}$ at time $\tau+1 $ and after for all agents efficiently, and  $\sum_{k=1}^{6} p_{k}=1$. With these interactive trajectories, we can easily get $s_{n=2,...,N} ^{T+\tau+1 } $ for other agents.

\subsection{Cost function}

After expansion and rollout, we have to comprehensively consider the route from efficiency and safety~\cite{dai2021towards}. Our cost is a linear function denoted as $c= \omega_{1}c^{(1)}+\omega_{2}c^{(2)}+\omega_{3}c^{(3)}+\omega_{4}c^{(4)}+\omega_{5}c^{(5)}$, accounting for efficiency, comfort and safety. To guarantee ego vehicle moving forward as fast as possible at the speed limit $v_{max} $, we set $c^{(1)}  = 1-\left ( \frac{v_{\tau} -v_{max}}{v_{\tau}}  \right )^{2}  $. We set $c^{(2)}  = \sum_{t=T+1}^{\tau } \left ( a_{t}-a_{t-1}  \right ) ^{2} $ and $c^{(3)}  = \sum_{t=T+1}^{\tau } \left ( \delta _{t}-\delta _{t-1}  \right ) ^{2} $ to encourage acceleration and steering smoothness and $c^{(4)}  = \sum_{t=T+1}^{\tau } \left ( \ln{(1+\exp \kappa(a_{t}-\alpha   ) )}+ \ln{(1+\exp -\kappa(a_{t}-\beta    ) )} \right ) $ to penalize hard acceleration and braking, where $\alpha = 4\mathrm{m/s^{2} } $, $\beta  = -5\mathrm{m/s^{2} } $ and $\kappa =15$. To calculate the collision penalty between pairwise nearby vehicles, we assume these shapes are rectangles. Then we have reason to set $c^{(5)} =\sum_{t=T+1}^{\tau }\left[S\left(\lambda _{x}  (\Delta x_{t}+l_x)\right)+S\left(\lambda _{x}  (l_x-\Delta x_{t})\right)\right] \\
\cdot\left[S\left(\lambda _{y}  (\Delta y_{t}+l_y)\right)+S\left(\lambda _{y} (l_y-\Delta y_{t})\right)\right] 
$, where $\Delta x_{t}$ and $\Delta y_{t}$ represent the minimum distance from all the surrounding agents to the ego vehicle. This collision risk premium related parameters are set as $S\left ( \chi  \right )  = Sigmoid\left ( \chi  \right ) -\frac{1}{2} =\frac{1}{1+\exp (-\chi )} -\frac{1}{2}$ with $l_{x}=10.0\mathrm{m } $ and $l_{y}=2.0\mathrm{m } $. Besides, $\lambda _{x}=0.5$ and $\lambda _{y}=9.0$ are set to control the longitudinal and lateral range. Finally the corresponding weights are set by convention with $\omega_{1}=1.0$, $\omega_{2}=-0.01$, $\omega_{3}=-1.5$, $\omega_{4}=-1.0$ and $\omega_{5}=-14.0$.

\section{Experiments}
\label{sec:experiments}

\begin{table*}[t]
\centering
\caption{Ablation study results of four interactive scenarios.}
\label{comparison}
\resizebox{\textwidth}{!}{
\setlength{\tabcolsep}{2.0mm}{
\begin{tabular}{c|ccc|ccc|ccc|ccc}
\toprule
& \multicolumn{3}{c|}{Scene 1} & \multicolumn{3}{c|}{Scene 2} & \multicolumn{3}{c|}{Scene 3} & \multicolumn{3}{c}{Scene 4} \\
                        & C.T. (s) & A.V. (m/s)  & C.D. (m) & C.T. (s) & A.V. (m/s)  & C.D. (m) & C.T. (m) & A.V. (m/s)  & C.D. (m) & C.T. (s) & A.V. (m/s)  & C.D. (m) \\
\midrule
PS-Rule                 & 4.33 & 12.81 & 5.71 & 2.33 & 9.60  & 0.45 & 2.16 & 10.22 & 1.57 & 3.46 & 7.83  & 0.77 \\
PS-Niter                & 5.27 & 10.04 & 4.94 & 2.69 & 6.99  & 0.97 & 2.33 & 7.23  & 1.94 & 4.06 & 5.33  & 2.06 \\
PS-Fixed                & 3.62 & 17.81 & 6.07 & 2.44 & 10.60 & 0.37 & 1.96 & 13.49 & 3.22 & 3.13 & 9.41  & 0.36 \\
\textbf{PS }                     & \textbf{5.01} & \textbf{10.80} & \textbf{5.24} & \textbf{3.99} & \textbf{2.88}  & \textbf{4.22} & \textbf{2.22} & \textbf{7.38}  & \textbf{1.97} & \textbf{4.44} & \textbf{4.91}  & \textbf{2.22} \\
\bottomrule
\end{tabular}
}}
\end{table*}

\begin{figure*}[t]
  \centering
  \begin{minipage}[b]{0.23\textwidth}
    \centering
    \includegraphics[width=\textwidth]{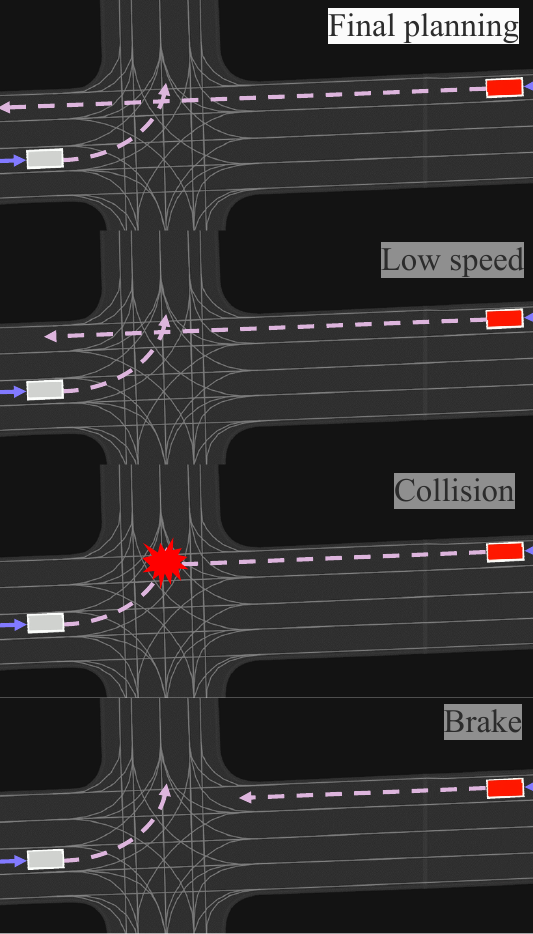}
    \caption*{(a)}
    \label{fig:a}
  \end{minipage}%
  \hspace{0.01\textwidth}
  \begin{minipage}[b]{0.23\textwidth}
    \centering
    \includegraphics[width=\textwidth]{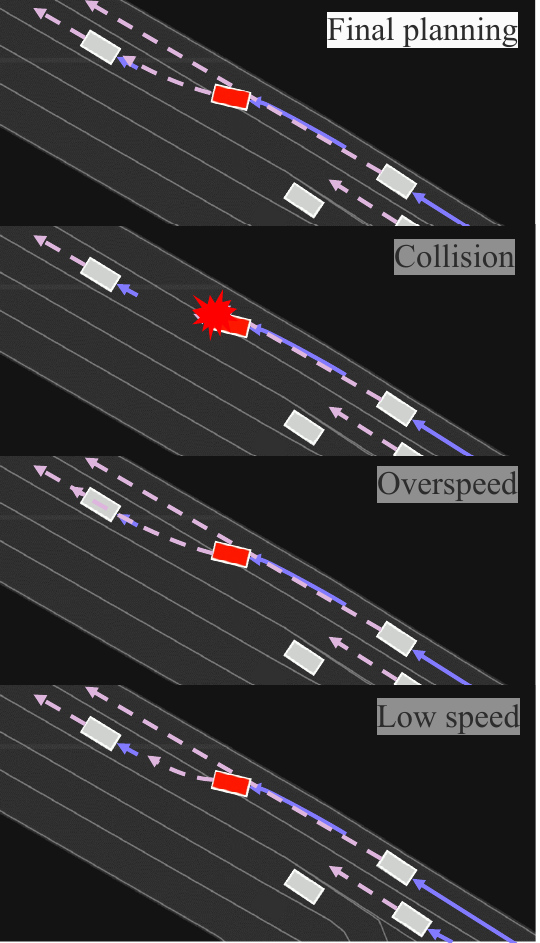}
    \caption*{(b)}
    \label{fig:b}
  \end{minipage}%
  \hspace{0.008\textwidth}
  \begin{minipage}[b]{0.23\textwidth}
    \centering
    \includegraphics[width=\textwidth]{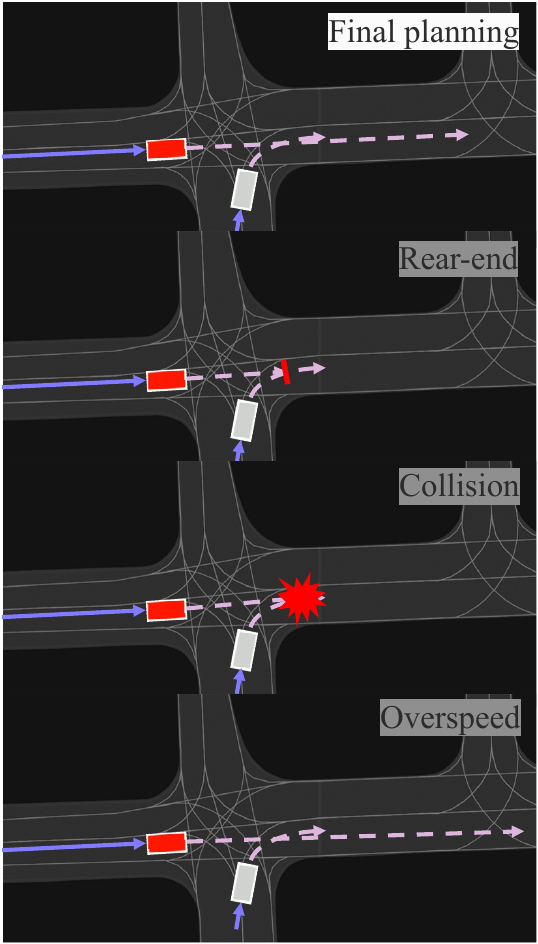}
    \caption*{(c)}
    \label{fig:c}
  \end{minipage}%
  \hspace{0.01\textwidth}
  \begin{minipage}[b]{0.23\textwidth}
    \centering
    \includegraphics[width=\textwidth]{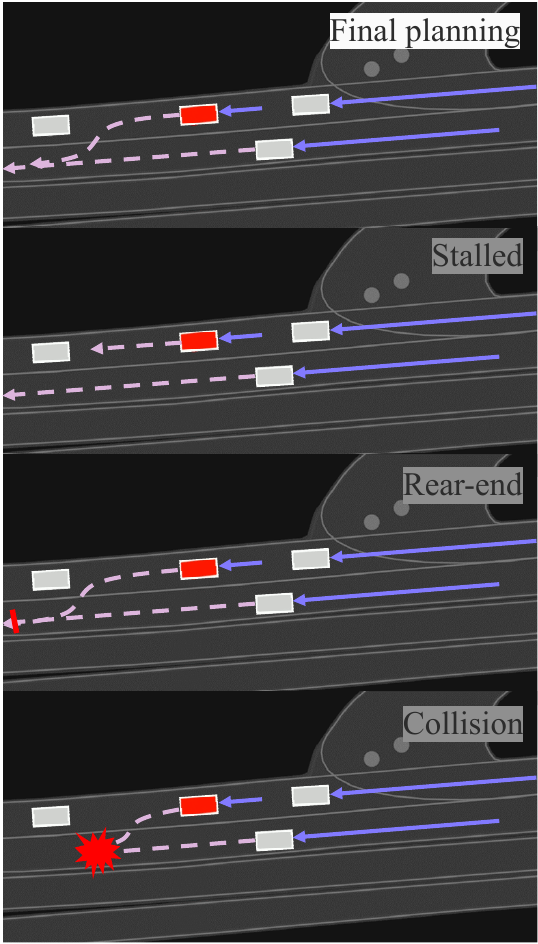}
    \caption*{(d)}
    \label{fig:d}
  \end{minipage}
  \caption{Test results and three counter-examples on four scenarios from Argoverse 2. The red bounding box represents the ego vehicle, and the grey bounding box represents the target agents. Purple curves represent history trajectories, and pink curves represent the planned trajectory of the ego vehicle and the actual future trajectories of target agents. Red stars and short red lines represent collisions and rear-ends occurring separately.}
  \vspace{-0.5cm}
  \label{fig:simulation}
\end{figure*}
\begin{table}
\centering
\caption{Comparison results of interactive scenarios.}
\label{comparison0}
\resizebox{\columnwidth}{!}{ 
\setlength{\tabcolsep}{2.0mm}{
\begin{tabular}{c|cccc}
\toprule
& \multicolumn{4}{c}{Average}    \\
                        & C.T. (s) & A.V. (m/s)  & C.D. (m)&P.E. (m) \\
\midrule
PS-MAE~\cite{cheng2023forecast}& 5.04& 12.80 & 3.22&3.78 \\
PS-FJMP~\cite{rowe2023fjmp}& 4.19& 15.4 & 2.98&4.22 \\

\textbf{PS }                     & \textbf{3.91}& \textbf{16.49}& \textbf{3.41}&\textbf{2.35}\\
\bottomrule
\end{tabular}
}}
\end{table}

\subsection{Implementation Details}
We evaluate our method in simulations of diverse urban driving scenarios from Argoverse 2~\cite{wilson2023argoverse}, with a 5-second observation window and a 6-second planning horizon at 10 Hz. The goal is to assess our model's planning in real-world scenarios. 
In the simulation, dynamic agents potentially interact with each other. Our route planner aims to find an appropriate route for the ego-vehicle to execute under 5 seconds agents' trajectories and lane polygons. 
All the experiments are conducted on a desktop equipped with an Intel I7-8700K CPU and a single  NVIDIA RTX 4090.

\subsection{Scenarios and Results}
To verify that our proposed method can automatically adapt to different traffic conditions with different semantic information, we present our planner results of four representative test cases and what will happen if ego vehicle does not operate our actions.
1) Vehicles in the opposing lanes make a left turn. As illustrated in \autoref{fig:simulation}-(a), it is used to verify the capability of dynamically adjusting speed changes facing uncertain hinder and keeping among the speed limit at the same time. We can see that improper speed change may cause a collision or a hard brake.
2) The ego vehicle initiates a left lane change into the gap between two adjacent vehicles. This case is to validate the capability of dealing with rapidly approaching vehicles from the rear and rear-side. As illustrated in \autoref{fig:simulation}-(b), our method conducts a safe and smooth lane change without collision with sluggish front care and aggressive rear vehicle. 
3) Target vehicle at the intersection merges in. This scenario shows that our method plans efficient trajectories when the lateral vehicle at the intersection tries to merge into the traffic flow. As shown in \autoref{fig:simulation}-(c), the inadequate plan may cause a rear end or exceed the speed limit.
4) Ego vehicle overtakes on crowded road. This case is used to verify the capability of quickly responding to complex interactions with other agents during traffic negotiation. As shown in \autoref{fig:simulation}-(d), when the front vehicle suddenly stops, and the ego vehicle has to change lanes with a rapidly moving vehicle, our method efficiently finds safe and feasible trajectories.

\subsection{Comparisons}

We compared the following algorithms in scenarios S1–S4. To highlight the concept of query-centric trajectory modeling, we employed a rule-based prediction approach, specifically utilizing a constant velocity lane-following strategy. This method is referred to as PS-Rule in the table. To prove the necessity of cyclic utilization between prediction and planning, we tested only inferring prediction once for each action, named as PS-Niter. Specifically, we applied a fixed action set rather than our adaptive action generation in the third ablation study, denoted as PS-Fix in the table.

We use three metrics to evaluate the performance of these methods: completion Time, average velocity and collision distance. Completion Time and average velocity relate to efficiency. They evaluate how fast we reach a target while complying with some speed limitations. Collision distance shows the minimized distance between surrounding agents and the ego vehicle. 
These three indexes are abbreviated as C.T., A.V., and C.D., respectively, in \autoref{comparison}.

In Scenario 1, while the fixed action set method aims to enable the vehicle to reach the target point more efficiently, it results in an average speed exceeding the 15 m/s limit. In Scenarios 2 and 4, certain ablation methods lead to collisions. For Scenario 3, both the rule-based prediction method and the fixed action set method may cause abrupt acceleration.

Utilizing prediction methods Forecast-MAE~\cite{cheng2023forecast} and FJMP~\cite{rowe2023fjmp} from Argoverse 2, we conducted evaluations across the four aforementioned scenarios. Extra metrics Planning Error (P.E.) from actual planning trajectory is calculated. The results are averaged, yielding the data presented in \autoref{comparison0}. We can see that PS has faster completion time than planners with prediction that output trajectories directly.

\section{CONCLUSIONS}
\label{sec:conclusion}
In this paper, we present a trajectory planning system, PS, that integrates planning and prediction by inferring potential trajectories of surrounding agents through an Monte Carlo search tree. The system continuously predicts future trajectories by feeding ego vehicle plans into the search tree, enabling dynamic trajectory exploration. The use of a Frenét frame-based ego planner and query-driven trajectory encoding ensures efficient and feasible nodes expansion within the search tree. Our proposed approach supports parallel scene simulations with reasonable actions. Key future research
 directions include developing cost learning mechanisms to
 replace the fixed cost function and adopting trajectory prediction based on continuous action sets.




\newpage
\bibliographystyle{plain}
\bibliography{reference}

\end{document}